\pdfoutput=1

\documentclass[11pt]{article}

\usepackage[]{acl}

\usepackage{times}
\usepackage{latexsym}

\usepackage[T1]{fontenc}

\usepackage[utf8]{inputenc}

\usepackage{microtype}

\usepackage{inconsolata}

\usepackage{comment}
\usepackage{xcolor}
\usepackage{booktabs}
\usepackage{graphicx}
\usepackage[singlelinecheck=false ]{caption}
\usepackage{amsmath}
\usepackage{cleveref}

\newcommand\blfootnote[1]{%
  \begingroup
  \renewcommand\thefootnote{}\footnote{#1}%
  \addtocounter{footnote}{-1}%
  \endgroup
}

\title{On the Role of Summary Content Units in Text Summarization Evaluation}

\author{
\begin{minipage}[t]{\textwidth}
\centering
\normalsize
Marcel Nawrath$^{1}$\textbf{*}, 
Agnieszka Nowak$^{1}$\textbf{*}, 
Tristan Ratz$^{1}$\textbf{*}, 
Danilo C. Walenta$^{1}$\textbf{*},
Juri Opitz$^{2}$\textbf{*},
Leonardo F. R. Ribeiro$^{3}$\textbf{$\dagger$},
João Sedoc$^{4}$,
Daniel Deutsch$^{5}$,
Simon Mille$^{6}$,
Yixin Liu$^{7}$,
Lining Zhang$^{4}$,
Sebastian Gehrmann$^{8}$,
Saad Mahamood$^{9}$,
Miruna Clinciu$^{10}$,
Khyathi Chandu$^{11}$,
Yufang Hou$^{12}$\textbf{*$\ddagger$} \\
{\footnotesize \normalfont 
$^{1}$TU Darmstadt,
$^{2}$University of Zurich,
$^{3}$Amazon AGI,
$^{4}$New York University,
$^{5}$Google Research,
$^{6}$ADAPT Centre, DCU,
$^{7}$Yale University,
$^{8}$Bloomberg,
$^{9}$trivago N.V.,
$^{10}$University of Edinburgh,
$^{11}$Allen Institute for AI,
$^{12}$IBM Research Europe, Ireland
} 
\end{minipage}
}

\begin{document}
\maketitle
\blfootnote{\textbf{*} Equal contributions.}
\blfootnote{\textbf{$\dagger$} Work done prior to joining Amazon.}
\blfootnote{\textbf{$\ddagger$} Correspondence to yhou@ie.ibm.com.}

\begin{abstract}
At the heart of the Pyramid evaluation method for text summarization lie human written summary content units (SCUs). These SCUs are concise sentences that decompose a summary into small facts. Such SCUs can be used to judge the quality of a candidate summary, possibly partially automated via natural language inference (NLI) systems. Interestingly, with the aim to fully automate the Pyramid evaluation, \newcite{zhang-bansal-2021-finding} show that SCUs can be approximated by automatically generated semantic role triplets (STUs). However, several questions currently lack answers, in particular: i) Are there other ways of approximating SCUs that can offer advantages? ii) Under which conditions are SCUs (or their approximations) offering the most value? In this work, we examine two novel strategies to approximate SCUs: generating SCU approximations from AMR meaning representations (SMUs) and from large language models (SGUs), respectively. We find that while STUs and SMUs are competitive, the best approximation quality is achieved by SGUs. We also show through a simple sentence-decomposition baseline (SSUs) that SCUs (and their approximations) offer the most value when ranking short summaries, but may not help as much when ranking systems or longer summaries.

\end{abstract} 

\section{Introduction}

Judging the quality of a summary is a challenging task. Besides being short and faithful to its source document, a summary should particularly excel in \textit{relevance}, that is, a summary should select only the most relevant or salient facts from a source document. An attractive method for assessing such notion of relevance is the \textit{Pyramid}-method \cite{nenkova-passonneau-2004-evaluating} that is based on so-called \textit{Summary Content Units} (SCUs) which decompose a reference summary into concise human-written English sentences. With SCUs available from one or different reference summaries, we can then more objectively assess the degree to which a generated summary contains the relevant information. With the aim to fully automate the Pyramid method, \citet{zhang-bansal-2021-finding} suggest that the required human effort can be partially and even fully alleviated, by i) automatically generating SCUs and ii) validating the relevance of a generated summary with a natural language inference (NLI) system that checks how many SCUs are entailed by the generated summary. 

Since strong NLI systems are available off-the-shelf and are known to be useful in natural language generation (NLG) evaluation \cite{chen2022menli,steen2023little}, the generation of SCUs appears as the most challenging and least-understood part of an automated Pyramid method. Indeed, while \citet{zhang-bansal-2021-finding} show that SCUs can be approximated by phrasing semantic role triplets using a semantic role labeler and coreference resolver, we lack availability and understanding of possible alternatives as well as their potential impact on downstream-task summary evaluation in different scenarios. 

In this work, we proposed two novel approaches to approximate SCUs: \emph{semantic meaning units} (SMUs) that are based on abstract meaning representation (AMR) and \emph{semantic GPT units} (SGUs) that leverage SoTA large language models (LLMs). We carry out experiments to systematically evaluate the intrinsic quality of SCUs and their approximations. On the downstream task evaluation, we find that although SCUs remain the most effective metric to rank different systems or generated summaries across three meta-evaluation datasets, surprisingly, an efficient sentence-splitting baseline already yields competitive results when compared to SCUs. In fact, the sentence-splitting baseline outperforms the best SCU approximation method on a few datasets when ranking systems or long generated summaries. 

In summary, our work provides important insights into the application of automation of the Pyramid method in different scenarios for evaluating generated summaries. We make the code publicly available at \url{https://github.com/tristanratz/SCU-text-evaluation/}.

\section{Related work}

Over the past two decades, researchers have proposed a wide range of human-in-the-loop or automatic metrics to assess the quality of generated summaries in different dimensions, including linguistic quality, coherence, faithfulness, and content quality. For more in-depth surveys on this topic, please refer to  \newcite{howcroft-etal-2020-twenty}, \newcite{ELKASSAS2021113679}, and \newcite{gehrmann2022repairing}.

In this work, we focus on evaluating the content quality of a generated summary that assesses whether the summary effectively captures the salient information of interest from the input document(s). In the reference-based metrics, content quality is assessed by comparing system-generated summaries to human-written reference summaries. The Pyramid method \cite{nenkova-passonneau-2004-evaluating} is regarded as a reliable and objective approach for assessing the content quality of a generated system summary. Below we provide a brief overview of the Pyramid method and highlight several previous efforts to automate this process.

\paragraph{Pyramid Method.} The original Pyramid method \cite{nenkova-passonneau-2004-evaluating} comprises two steps: SCUs creation and system evaluation. 
In the first step, human annotators exhaustively identify Summary Content Units (SCUs) from the reference summaries. Each SCU is a concise sentence or phrase that describes a single fact. The weight of an SCU is determined by the number of references in which it occurs. In the second step, the presence of each SCU in a system summary is manually checked. The system summary's Pyramid score is calculated as the normalized sum of the weights of the SCUs that are present. Later, \newcite{shapira-etal-2019-crowdsourcing} introduce a revised version of the original Pyramid method where they eliminate the merging and weighting of SCUs, thereby enabling SCUs with the same meaning to coexist.

\paragraph{Automation of the Pyramid Method.} Given the high cost and expertise required for implementing the Pyramid method, there have been attempts to automate this approach in recent years. \newcite{peyrard-etal-2017-learning} propose an automatically learned metric to directly predict human Pyramid scores based on a set of features. \newcite{zhang-bansal-2021-finding} propose a system called \emph{$Lite^{3}Pyramid$} that uses a semantic role labeler to extract semantic triplet units (STUs) to approximate SCUs. They further use a trained natural language inference (NLI) model to replace the manual work of assessing SCUs' presence in system summaries. In our work, we explore two new methods to approximate SCUs. We further investigate the effectiveness of the automated Pyramid method in different scenarios.

\section{SCU approximation I: SMU from AMR} 

\emph{Abstract Meaning Representation} (AMR) \cite{banarescu2013abstract} is a widely-used semantic formalism employed to encode the meaning of natural language text in the form of rooted, directed, edge-labeled, and leaf-labeled graphs. The AMR graph structure facilitates machine-readable explicit representations of textual meaning.

Motivated by \newcite{zhang-bansal-2021-finding}'s observation that STUs based on semantic roles cannot well present single facts in long reference summary sentences that contain a lot of modifiers, adverbial phrases, or complements, we hypothesize that AMR has the potential to capture such factual information more effectively. This is because, in addition to capturing semantic roles,  AMR models finer nuances of semantics, including negations, inverse semantic relations, and coreference. 

To generate semantic meaning units (SMUs) from a reference summary, we employ a \textbf{parser} that projects each sentence of the reference summary onto an AMR graph, then split the AMR graph into meaningful, event-oriented subgraphs.
Finally, we use a \textbf{generator} to generate a text (an SCU approximation) from each subgraph.\footnote{For  \textbf{parser} and \textbf{generator}, we use strong off-the-shelf models from \url{https://github.com/bjascob/amrlib-models}: \emph{parse\_xfm\_bart\_large} and \emph{generate\_t5wtense}. \emph{parse\_xfm\_bart\_large} is fine-tuned on \emph{BART\_large}. The model achieves a high Smatch score on the standard AMR benchmark (83.7 SMATCH on the AMR-3 test dataset).} 

\begin{figure}[t]
    \centering
    \includegraphics[width=0.48\textwidth]{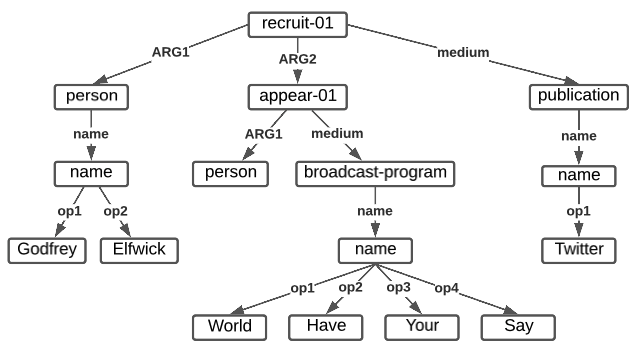} 
    \caption{The AMR graph for the sentence ``\emph{Godfrey Elfwick recruited via Twitter to appear on World Have Your Say}''}
    \label{fig:amrtree}
\end{figure}

\begin{figure}[t]
    \centering
    \includegraphics[width=0.48\textwidth]{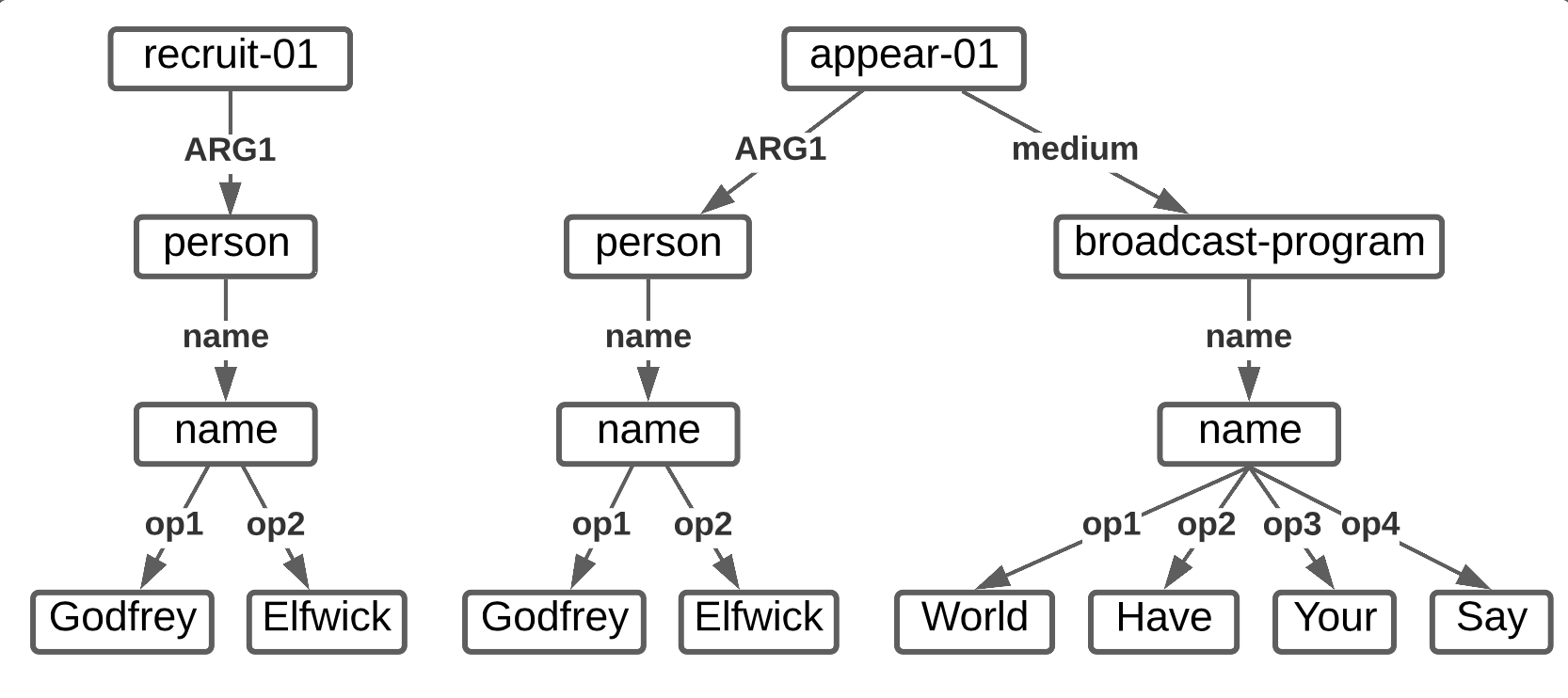} 
    \caption{Two AMR sub-graphs for the sentence ``\emph{Godfrey Elfwick recruited via Twitter to appear on World Have Your Say}''}
    \label{fig:subtrees}
\end{figure}

While there may be various conceivable ways to \textit{extract subgraphs}, for our experiment we use simple and intuitive \textbf{splitting }heuristics. Given an AMR graph, we first extract all predicates to discern their semantic meaning as we view them to form the core of a sentence's meaning. Subsequently, the argument connections within the predicates were examined. If a predicate is connected to at least one core role (CR), indicated by \texttt{ARG$_n$} edge labels, we extract a sub-graph for every CR of this predicate containing the CR and the underlying connections. Figure \ref{fig:subtrees} shows an example of two extracted subgraphs from the AMR graph in Figure \ref{fig:amrtree}
for the input sentence ``\textit{Godfrey Elfwick recruited via Twitter to appear on World Have Your Say''} based on the \textbf{parser} and \textbf{splitting} steps. 
Finally, we generate two SMUs by applying the \textbf{generator} to the two subgraphs in Figure \ref{fig:subtrees}:
\begin{itemize}
    \item SMU 1: \textit{Godfrey Elfwick was recruited.} 
    \item SMU 2:  \textit{Godfrey Elfwick will appear on World Have Your Say.}
\end{itemize}

\section{SCU approximation II: SGU from LLM} Recently, it became widely known that pre-trained large language models (LLMs) are able to generate high-quality output according to prompts given by humans, optionally exploiting a few shown examples through in-context learning \cite{brown2020language}. Therefore, we try to approximate SCUs using GPT models from OpenAI, calling the resulting units Semantic GPT Units (SGUs). Specifically, we use GPT-3.5-Turbo which is built on InstructGPT \cite{ouyang2022training} and GPT-4 \cite{openai2023gpt4} to generate SGUs (SGUs\_3.5 and SGUs\_4) for each reference summary using the same prompt and a one-shot example. Please refer to Appendix \ref{SGUDetails} for more details.

\section{Experiments}

\subsection{Dataset and NLI models}
\paragraph{Datasets.} We run the experiments on four existing English meta-evaluation datasets:
(1) TAC08 \citep{tac08}, (2) TAC09 \citep{tac09}, (3) REALSumm \citep{bhandari-etal-2020-evaluating} and (4) PyrXSum \citep{zhang-bansal-2021-finding}. We use TAC08 for development purposes and evaluate the results on the last three datasets. Each dataset contains one or multiple reference summaries, the corresponding human-written SCUs, the generated summaries from different systems, and the human evaluation result for each summary/system based on the Pyramid method. Table \ref{tab:dataset} shows some statistics of the reference summaries across different datasets. In general, PyrXSum contains short and abstractive summaries, while RealSumm and TAC09 contain long and extractive summaries. More details on the datasets can be found in \cref{sec:appendixData}.

\paragraph{NLI Models.} We use the NLI model from \newcite{zhang-bansal-2021-finding}\footnote{The model can be downloaded from the HF model hub: https://huggingface.co/shiyue/roberta-large-tac08.} that was fine-tuned on TAC08's SCU presence gold annotations based on a NLI model from \newcite{nie-etal-2020-adversarial}. Following \newcite{zhang-bansal-2021-finding}, we use the fine-tuned NLI model with the probability of presence label to calculate the Pyramid score of a generated summary. Please refer to Appendix \ref{sec:appendixLevel} for more details.

\subsection{Baselines}
\paragraph{STUs} are short sentences that are based on semantic role (SR) triples from an SR-labeling and coreference system \cite{zhang-bansal-2021-finding}.

\paragraph{Sentence splitting} is a baseline that may shed light on the overall usefulness of SCUs in summary evaluation. We split every reference summary into sentences and treat them as SCU approximations.

\paragraph{N-grams} consist of phrases randomly extracted from a reference summary. For each sentence from the summary, we naively generate all possible combinations of $\{3, 4, 5\}$ consecutive words. We randomly select a subset from these combinations, which accounts for 5\% of all combinations.

\subsection{Intrinsic Evaluation}
As proposed by \citet{zhang-bansal-2021-finding}, we evaluate approximation quality with an \textit{easiness score}. The score is built by iterating over each SCU-SxU pair and averaging over the maximum ROUGE-1-F1 score found for each SCU. Naturally this score is recall-biased, and therefore, we also present the score calculated in the reverse direction, to evaluate the precision of our approximated SCUs (c.f.\ Appendix \ref{sec:intrinsicDetails} for more details).
The results are shown in Table \ref{tab:intrinsicEval}. We find that best approximation quality for RealSumm is achieved by STUs, while for PyrXSum, SGU\_4 performs best. Considering the longer texts of TAC09, STUs excel in recall, while SGU\_4 excels in precision.

\begin{table}[t]
\centering
\scalebox{0.9}{\begin{tabular}{l l l l}
\toprule
 & RealSumm & PyrXSum & TAC09 \\  \hline
 Avg. \# sent. & 4.73& 2.02& 27.22 \\ \hline
Avg. \# words & 63.71 & 20.56 & 386.82 \\ \hline
Avg. \# words/sent & 13.47 & 10.18 & 14.21 \\ \hline
\# ref summary& 1& 1& 4 \\ \hline 
Avg. \# SCUs& 10.56 & 4.78 & 31.63 \\ 

\bottomrule
\end{tabular}}

\caption{Statistics of the reference summaries from different datasets.
}
\label{tab:dataset}
\end{table}

\begin{table}[t]
\centering
\scalebox{0.9}{\begin{tabular}{l ll ll ll}
\toprule
 & \multicolumn{2}{c}{RealSumm} & \multicolumn{2}{c}{PyrXSum} & \multicolumn{2}{c}{TAC09} \\ \cline{2-7} 
Metrics & R & P & R & P & R & P \\ \hline
sentence split & .54 & .67 & .41 & .54 & .50 & .54 \\
ngrams & .41 & .52 & .38 & .52 & .46 & .39 \\ \hline
STUs & \textbf{.66} & .68 & .54 & .65 & \textbf{.61} & .53 \\ \hline
SMUs & .56 & .58 & .53 & .58 & .52 & .48 \\ \hline
SGUs\_3.5 & .58 & .67 &.58  & .63 &  .36& .48 \\
SGUs\_4 & .61 & \textbf{.69} & \textbf{.61} & \textbf{.66} & .52 & \textbf{.61} \\ 
\bottomrule
\end{tabular}}

\caption{Intrinsic evaluation results. R is the recall-oriented simulation easiness score from \newcite{zhang-bansal-2021-finding}, while P is our precision-oriented score that is computed in the reverse direction.}
\label{tab:intrinsicEval}
\end{table}

\begin{table*}[ht]
\begin{small}
\centering
\resizebox{\textwidth}{!}{
\begin{tabular}{l lll lll l lll lll}
\toprule
 & \multicolumn{6}{c}{System-Level} &  & \multicolumn{6}{c}{Summary-Level}   \\ \cline{2-7} \cline{9-14}
 &
 
  \multicolumn{2}{l}{RealSumm} &
  \multicolumn{2}{l}{PyrXSum} &
  \multicolumn{2}{l}{TAC09} &
   &
 
  \multicolumn{2}{l}{RealSumm} &
  \multicolumn{2}{l}{PyrXSum} &
  \multicolumn{2}{l}{TAC09} \\
Metrics        & $r$ & $\rho$ & $r$ & $\rho$ & $r$ & $\rho$ &  & $r$ & $\rho$  & $r$ & $\rho$ & $r$   & $\rho$     \\ \hline

SCUs &.95& .95& .98& .98& .99& .97&& .59& .58& .70& .69&  .76& .70\\
\hline
SCU Approximations && & & & & && & & & &&\\ \hline
- sentence split & .93& \textbf{.95}& \textbf{.97}& \textbf{.97}&  .97& .94&& .48& .46& .37& .36&  \textbf{.73}& .66\\
    
   - ngrams & .90& .92& .94& .82&  .96& .92&& .36& .35& .38& .38&  .65& .61\\
    \hline
   - STUs & .92& .94& .95& .95&  \textbf{.98}& .95&& .51& .50& .46& .44&  \textbf{.73}& \textbf{.67}\\
    \hline
   - SMUs & \textbf{.94}& .94& .96& .94&  \textbf{.98}& \textbf{.96}&& .50& .48& .46& .44& .70& .64\\
    \hline
   - SGUs\_3.5 & .93& \textbf{.95}& \textbf{.97}& .93&  .96& .88&& .49& .46& .56& .55& .54& .49\\
   - SGUs\_4 & .92& .94& \textbf{.97}& .95& \textbf{.98}& \textbf{.96}&& \textbf{.54}& \textbf{.52}& \textbf{.58}& \textbf{.56}& .71& .66\\
\bottomrule
\end{tabular}}
\caption{Results of different metrics on three datasets. Best numbers among all SCU approximations are bolded.}
\label{tab:results}
\end{small}
\end{table*}

\subsection{Extrinsic Evaluation}
Our downstream evaluation consists of two parts: summary quality evaluation at the system and summary levels, respectively. System-level correlation assessment evaluates the ability of the metric to compare different summary systems individually.
In contrast, summary-level evaluation determines the metric's ability to compare summaries created by different systems for a common set of documents.
Following \newcite{zhang-bansal-2021-finding}, we use Pearson $r$ and Spearman $\rho$ to evaluate the correlations between metrics with gold human labeling scores. Pearson measures linear correlation and Spearman measures ranking correlation. Please refer to Appendix \ref{sec:appendixLevel} for more details about using the NLI model to score a generated summary and how to calculate these two types of correlations.

The results are shown in Table \ref{tab:results}.\footnote{Note that we do not include comparisons with the recent automatic evaluation metrics based on LLMs such as GPTScore \cite{fu2023gptscore}.  Recent studies pointed out that these automatic metrics are not as effective as the traditional automatic evaluation metrics, such as ROUGE-1, to compare the summaries of different systems in terms of content coverage \cite{liu-etal-2023-revisiting}.} In general, SGUs offer the most useful SCU approximation, with the exception of  TAC09 (summary-level), where STUs remain the best approximation method, slightly outperforming our simple sentence-splitting baseline. However, SGUs still lack the usefulness of true SCUs, which seem to remain the most useful way to evaluate system summary quality (if resources permit). Interestingly, however, to discriminate the quality of systems, it is enough to use any approximation, even the sentence split baseline is sufficient to accurately discriminate between systems.

\subsection{Human Evaluation}

For a representative sample of human results of our experiment, three authors evaluated the quality of SCUs, STUs, SMUs and SGUs\_4  for 10 reference summaries randomly sampled from REALSumm and PyrXSum, annotating each of 40 examples according to 3 dimensions: Well-formedness, Descriptiveness and Absence of hallucination, amounting to a total of 240 annotation hits. Please refer to  Appendix \ref{sec:humaneval} for more details of the annotation scheme. 

Overall, Cohen’s $\kappa$ scores among three annotators range from 0.37 to 0.87. After a thorough check, we found that all annotators agree on the general trend (i.e., SCUs and SGUs are generally better than SMUs and STUs). One annotator appeared to diverge from the other two by slightly favoring SMUs over STUs. To increase the power of the experiment, two annotators then annotated another 20 summaries each, resulting in an additional 480 annotation hits.  

\begin{figure}[t]
\centering
\includegraphics[width=\linewidth]{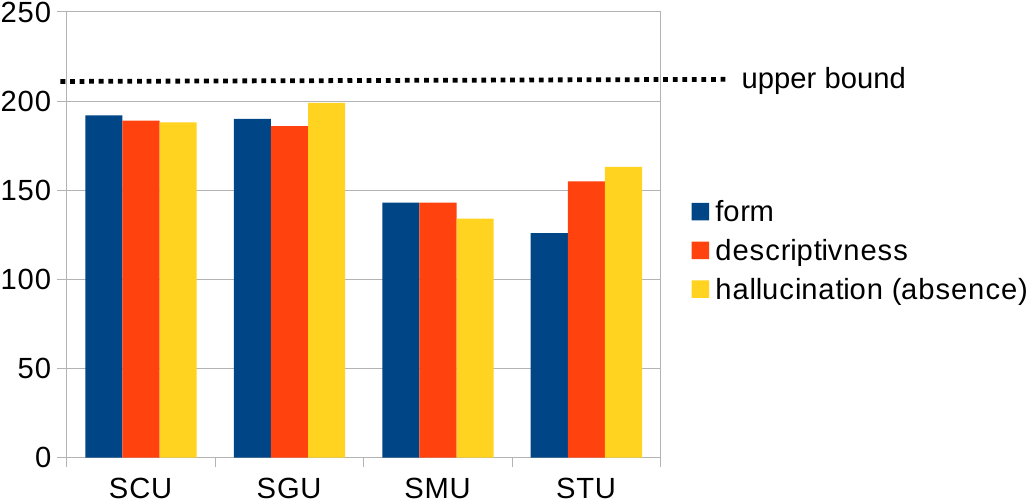}
    \caption{Human evaluation results. Each bar represents the sum of scores aggregated over all annotators. Upper-bound indicates the best possible result (each annotator always assigns the maximum quality score).}
    \label{fig:humaneval}
\end{figure}

The findings shown in Figure \ref{fig:humaneval} illustrate that the quality of SMUs is comparable to the STUs. But the revealed overall trend is clearly that SCUs (human written units) and SGUs (LLM generated units) achieve \textit{similar} and \textit{very high} quality, while STUs (triplet units based on SRL) and SMUs (units from AMR semantic graph) are similarly of lower quality. To see if these differences are significant, we calculate the Wilcoxon signed-rank test. For all categories (descriptiveness, well-formedness, and absence of hallucination), the human SCUs and SGUs are not of significantly different quality ($p < 0.005$). However, both SCUs and SGUs are of significantly better quality than STUs and SMUs ($p < 0.005$). Within SMUs and STUs, the categories of descriptiveness and well-formedness are not of significantly different quality ($p < 0.005$). However, STUs are significantly better in reduction of hallucination compared with SMUs ($p < 0.005$), an outcome that could be explained either by destruction of coherent information when splitting the AMR graph, or hallucination of the AMR models.\footnote{While AMR parsers nowadays achieve impressive scores on benchmarks \cite{graphene,lee-etal-2022-maximum}, recent research shows that they still make crucial errors \cite{opitz-frank-2022-better, groschwitz-etal-2023-amr}.}

{The result of the human annotation, however, must not be taken as proof that there is quality parity of SCUs and SGUs. Indeed, when contrasting the finding of the human evaluation, where SCUs and SGUs appear of similar high quality, against the empirical finding that SCUs provide substantially better downstream performance \textit{for shorter texts in summary-level evaluation}, we have reason to believe that there is a quality aspect of SCUs that both LLM/ChatGPT and our annotation setup failed to measure.

\section{Discussion and conclusions}
This work focuses on automating the Pyramid method by proposing and evaluating two new methods to approximate SCUs. We found out that there are more effective ways of approximating SCUs than with STUs only, and our extrinsic evaluation suggests that costly SCUs and approximations may even be unnecessary for system comparisons.

There are several aspects worth discussing. Firstly, as shown by comparing, e.g., the results of STUs and SMUs in Table~\ref{tab:intrinsicEval} and Figure~\ref{fig:humaneval}, it appears that ROUGE-1-F1 exhibits a weak correlation with human evaluation. This raises concerns about the effectiveness of using this metric in previous studies to evaluate the quality of SCU approximations. Secondly, it seems that we may not need the costly SCUs and their approximations to compare summarization systems or rank long generated summaries (TAC09). Surprisingly, a simple sentence splitting baseline already achieves competitive results compared to SCUs on these tasks, while automatically obtained SGUs generally score high both on system- and summary-level evaluations. Finally, SCUs and their approximations offer the most value for summary-level evaluation, especially when summaries are rather short (PyrXSum and RealSumm). 

\section*{Limitations}

First, we would have liked to  achieve better performance with SMUs generated from an AMR. In theory, using AMR graph splitting would ideally decompose a textual meaning into parts, and the AMR generation systems promise to phrase any such sub-graph into natural language.
Inspecting all three parts of our SMU generation pipeline (parsing, splitting, and generating), we find that some issues may be due to our manually designed splitting strategy being too naive. While the rules are simple and their creation has profited from communication with AMR-knowledgeable researchers, a main problem is that there are countless possibilities of how to split an AMR graph, and the importance of rules can depend on the graph context. Therefore, we believe that future work can strongly improve the AMR-based approach by learning how to better split meaning representation graphs.

Second, our NLI system was fine-tuned on gold SCUs extracted from the development data (TAC08), since this was found to work best by \newcite{zhang-bansal-2021-finding}. While in principle this does not affect the evaluation of SxUs, which was the focus of this paper, it is not unlikely that by training the NLI system on each SxU type separately, the results of SxUs may further improve. Therefore the results for human written SCUs can be considered slightly optimistic. In general, the interaction of NLI and SCUs in an automated Pyramid method needs to be better understood. Other recent findings \cite{chen2022menli,steen2023little} suggest that NLI models may play an underestimated role in NLG evaluation. As a check, we repeated the evaluation with an NLI system without SCU fine-tuning, and observe significant performance drops across the board, indicating that (i) SCU results are likely not too over-optimistic in comparison to SxUs,  and (ii) the effective adaptation strategy of the NLI system may be the second cornerstone of an accurate automatic Pyramid method and thus should be explored in future work. 

Finally, although our results offer insights into the design choices when applying the automatic Pyramid method for text summarization evaluation in different scenarios (short vs. long summaries), we do not explore its applications beyond summary and summarizer evaluation and ranking, such as using basic meaning units to evaluate the factual precision in long-form text generation \cite{min2023factscore}. We leave this for future studies.

\bibliography{main}

\appendix
\section{Appendix}
\label{sec:appendix}

\subsection{Obtaining SGUs from the GPT Models}
\label{SGUDetails}
Below we show an example of the prompt we use to obtain SGUs from the GPT-3.5-Turbo and GPT-4 models through OpenAI APIs\footnote{https://openai.com/blog/openai-api}. We randomly choose a summary and the corresponding human-written SCUs from the TAC08 dataset as the one-shot example.
We did not alter the standard parameters except for temperature which we set to 0 to ensure reproducibility.

 \paragraph{System Instruction} You split the provided input in small sentences separated by an \#. The split sentences represent subsentences of the original sentences. 
 
 \paragraph{Example Input} Irish PM Ahern said the main goal of the US-brokered Good Friday pact of 1998, a joint Catholic-Protestant administration in Northern Ireland, could be revived only with a complete end of IRA weapons use. The landmark peace deal led to a virtual end of violence in that area. Sinn Fein leader Gerry Adams has appealed to IRA members to end their armed struggle in favor of democratic politics. Hopes are rising in Northern Ireland that the IRA will disarm. British PM Blair and Ahern will chair a review of the Northern Ireland situation in London.
 
 \paragraph{Example Output} Good Friday pact was agreed in 1998 \# Good Friday pact was a peace pact \# Good Friday pact set up a joint Catholic-Protestant administration in Northern Ireland \# Good Friday pact was mediated by the US \# Irish Republican Army increased activity \# Irish PM Ahern called to end violence \# Sinn Fein Adams called to end violence \# Hope in Northern Ireland that the IRA will disarm \# British PM Blair and Ahern will chair a review of the Northern Ireland situation in London 

\paragraph{Testing Reference Summary} Netherlands midfielder Wesley Sneijder has joined French Ligue 1 side Nice on a free transfer. 

\paragraph{Output From GPT-3.5-Turbo} \emph{Netherlands midfielder Wesley Sneijder has joined Nice \# Sneijder was a free transfer \# Nice is a French Ligue 1 side} 

\paragraph{Output From GPT-4} \emph{Netherlands midfielder Wesley Sneijder \# Sneijder joined French Ligue 1 side Nice \# Joined on a free transfer}

\subsection{Dataset Details}
\label{sec:appendixData}

In general, all datasets (TAC08, TAC09, RealSumm, PyrXSum) contain: a) human written reference summaries; b) human expert written SCUs that are derived from the human written reference summaries; c) automatic summaries generated from different systems; d) SCU-presence labels for all system summaries that are labeled using either in-house annotators or Amazon Mechanical Turk (AMT). 

The TAC08 dataset includes 96 examples and outputs from 58 systems, while TAC09 contains 88 examples and outputs from 55 systems. Both datasets contain multiple reference summaries for each example, as well as the corresponding SCU annotations.

The REALSumm dataset contains 100 test examples from the CNN/DM dataset \citep{herrmann2015} and 25 system outputs. The SCUs are labeled by the authors and SCU-presence labels are collected using Amazon Mechanical Turk (AMT). 

PyrXSum \citep{zhang-bansal-2021-finding} includes 100 test examples from the XSum dataset \citep{narayan-etal-2018-dont}, which contains short and abstractive summaries.
Similar to REALSumm, the SCUs are manually labeled by the authors and SCU-presence labels are collected for summaries generated by 10 systems through AMT.
\\

\subsection{Intrinsic Evaluation Details}
\label{sec:intrinsicDetails}

We calculate two intrinsic evaluation metrics: A recall-based easiness score and a precision-based easiness score, denoted by $EasinessR$ and $EasinessP$. They evaluate how accurately the generated SxU units resemble human written SCUs. For a sentence with N human-written SCUs, 
$$
EasinessR = \frac{\sum Acc_j}{N},  
$$
where $$Acc_j = \max_m Rouge1_{F1}(SCU_j, SxU_m).$$
In the above formula, $Acc_j$ finds the SxU unit that has the max $Rouge1_{F1}$ score with $SCU_j$. $EasinessR$ corresponds to the easiness score defined in \newcite{zhang-bansal-2021-finding}.
To complement the recall-based easiness score, we introduce a precision-based $EasinessP$ that is calculated as:
$$
 EasinessP = \frac{\sum Acc_j}{N},  
$$
where $$Acc_j = \max_m Rouge1_{F1}(SxU_j, SCU_m).$$ This time, $Acc_j$ finds the SCU unit that has the max $Rouge1_{F1}$ score with $SxU_j$.

\subsection{Extrinsic Evaluation Details}
\label{sec:appendixLevel}
\paragraph{Details about NLI models} 
For extrinsic evaluation, we follow the previous method proposed in \newcite{zhang-bansal-2021-finding} and use the NLI model they fine-tuned on the TAC 2008 dataset. More specifically, based on the NLI model, a system summary $s$ will be scored as:
$$Score_s = \sum P_{NLI} (e|SxU_ j, s)/N,$$
where N is the total number of SxUs extracted from the gold reference summary or summaries, and $P_{NLI} (e|SxU_j, s)$ is the probability of the entailment class from the underlying NLI model that tells us how likely the unit $SxU_j$ is entailed by the system summary $s$. \newcite{zhang-bansal-2021-finding} explored different ways of using the NLI model, including a standard 3-class setting of the NLI model (entail/neutral/contradict) and a fine-tuned version of a 2-class setting (present/not-present), as well as using either the output probability of entailment/present class or the predicted 1 or 0 entailment/present label. They reported that using the fine-tuned model with the probability of the presence label works the best. We use this setup in our work.

\paragraph{Details about calculating correlations} 
{\emph{System-level}} correlation assesses the metric's ability to compare different summarization systems.
This is denoted as K and measures the correlation between human scores (h), the metric (m), and the generated summaries (s) for N examples across S systems in the meta-evaluation dataset. The system-level correlation is then defined as:

\begin{align}
K_{m,h}^{sys} = K(&[\frac{1}{N} \sum^{N}_{i=1} m(s_{i1}), ... ,\frac{1}{N} \sum^N_{i=1} m(s_{iS})],\nonumber\\
&[\frac{1}{N} \sum^{N}_{i=1} h(s_{i1}), ... ,\frac{1}{N} \sum^{N}_{i=1} h(s_{iS})])\nonumber
\end{align}

\noindent \emph{Summary-level} correlation assesses the metric's ability to compare summaries produced by different systems for a common document(s). The summary-level correlation is then defined as:

\begin{align}
K_{m,h}^{sum} = \frac{1}{N} \sum^{N}_{i=1}K(&[m(s_{i1}), ... , m(s_{iS})],\nonumber\\
&[h(s_{i1}), ... ,h(s_{iS})])\nonumber
\end{align}

\subsection{Human annotated evaluation}
\label{sec:humaneval}
The text units of each example were analyzed regarding Well-formedness, Descriptiveness and Absence of hallucination. For each dimension, we classified it into one of three categories based on the evaluator's satisfaction with the system's output. These categories ranged from ``1 - Unhappy with system output'', ``2 - implying dissatisfaction or a less than satisfactory result'', to ``3 - Okay with system output (3)''.
Below we denote ASCU for approximated summary content units (e.g., STUs, SMUs, and SGUs\_4) and SCUs. We provide a detailed definition for each evaluation category:

\begin{itemize}
    \item Well-formedness (surface quality)
    \begin{itemize}
        \item 1: Many ASCUs are not concise English sentences
        \item 2: Some ASCUs are not concise English sentences 
        \item 3: Almost all or all ASCUs are concise English sentences
    \end{itemize}
    \item Descriptiveness (meaning quality I)
    \begin{itemize}
        \item 1: Many meaning facts of the summary have not been captured well by the ASCUs
        \item 2: Some meaning facts of the summary have not been captured by the ASCUs
        \item 3: Almost every or every meaning fact of the summary has been captured by an ASCU
    \end{itemize}
    \item Absence of hallucination (meaning quality II)
    \begin{itemize}
        \item 1: Many ASCUs describe meaning that is not grounded in the summary
        \item 2: There is some amount of ASCUs that describe meaning that is not grounded in the summary
        \item 3: Almost no or no ASCU describes meaning that is not grounded in the summary 
    \end{itemize}
\end{itemize}

In the following, we show an example of the reference summary from PyrXSum and the corresponding SCUs and their approximations:

\begin{itemize}
    \item  \textbf{Reference summary:} West Ham say they are ``disappointed'' with a ruling that the terms of their rental of the Olympic Stadium from next season should be made public.
    \item \textbf{SCUs:} West Ham are ``disappointed'' with a ruling \#
The ruling is that their rental terms should be made public \# West Ham will rent the Olympic Stadium from next season

    \item \textbf{STUs:} West Ham say they are ``disappointed'' with a ruling that the terms of their rental of the Olympic Stadium from next season should be made public \#
They are ``disappointed'' with a ruling that the terms of their rental of the Olympic Stadium from next season should be made public \# should made public

    \item \textbf{SMUs:} West Ham say they are disappointed by the ruling that their terms of rental for the Olympic Stadium next season should be public \# The ruling that the terms of West Ham's Olympic Stadium rental next season should be public was disappointing \# West Ham rent the Olympic Stadium \#
West Ham will rent the Olympic Stadium next season

    \item \textbf{SGUs\_4:} West Ham is disappointed with a ruling \#
Terms of their Olympic Stadium rental should be made public \#
Olympic Stadium rental starts next season

\end{itemize}

\end{document}